\pdfoutput=1

\documentclass[11pt]{article}

\usepackage[final]{EMNLP2023}

\usepackage{times}
\usepackage{latexsym}

\usepackage[T1]{fontenc}

\usepackage[utf8]{inputenc}
\usepackage{hyperref}       
\usepackage{url}           
\usepackage{booktabs}      
\usepackage{amsfonts}      
\usepackage{nicefrac}  
\usepackage{xcolor}      
\usepackage{multirow}      
\usepackage{graphicx}       
\usepackage{tabularx}      
\usepackage{arydshln}       
\usepackage{adjustbox}
\usepackage{subcaption}     
\usepackage{marvosym}       
\usepackage{float}

\usepackage{microtype}

\usepackage{inconsolata}

\title{Towards Efficient Visual-Language Alignment \\of the Q-Former for Visual Reasoning Tasks}

\author{
Sungkyung Kim$^{1*}$ \quad Adam Lee$^{2*}$ \quad Junyoung Park$^3$ \\ \textbf{Andrew Chung$^{1,3}$} \quad \textbf{Jusang Oh$^1$} \quad \textbf{Jay-Yoon Lee}$^{4\dagger}$ \\
$^1$Seoul National University \quad $^2$UC Berkeley \quad $^3$Weavel, Inc. \\ $^4$Graduate School of Data Science, Seoul National University\\
\texttt{sk0428@snu.ac.kr, alee00@berkeley.edu} \\
\texttt{\{junyoung, sounho\}@weavel.ai}, 
\texttt{\{dhwntkd412, lee.jayyoon\}@snu.ac.kr}
}

\begin{document}
\maketitle

\renewcommand{\thefootnote}{\fnsymbol{footnote}} 
\footnotetext[1]{Equal contribution.}
\footnotetext[2]{Corresponding author.}
\begin{abstract}
Recent advancements in large language models have demonstrated enhanced capabilities in visual reasoning tasks by employing additional encoders for aligning different modalities. While the Q-Former has been widely used as a general encoder for aligning several modalities including image, video, audio, and 3D with large language models, previous works on its efficient training and the analysis of its individual components have been limited. In this work, we investigate the effectiveness of parameter efficient fine-tuning (PEFT) the Q-Former using InstructBLIP with visual reasoning benchmarks ScienceQA and IconQA. We observe that applying PEFT to the Q-Former achieves comparable performance to full fine-tuning using under 2\% of the trainable parameters. Additionally, we employ AdaLoRA for dynamic parameter budget reallocation to examine the relative importance of the Q-Former's sublayers with 4 different benchmarks. Our findings reveal that the self-attention layers are noticeably more important in perceptual visual-language reasoning tasks, and relative importance of FFN layers depends on the complexity of visual-language patterns involved in tasks. The code is available at \url{https://github.com/AttentionX/InstructBLIP\_PEFT}.
\end{abstract}

\section{Introduction}
Pre-trained large language models (LLMs) can be fine-tuned with instruction tuning to align the model responses with human intentions \cite{alpaca, wang2023selfinstruct}. Recently, model alignment with instruction tuning has been extended to the image domain by using an external encoder to align visual-language modalities and enhance the model's capabilities for visual reasoning. LLaVA \cite{liu2023visual,liu2023improved} uses a projection layer to project CLIP \cite{radford2021learning} image embeddings to the text embedding space of language models. However using a projection layer to convert every CLIP embedding from an image can take up a lot of context tokens and increase inference time. BLIP-2 \cite{li2023blip} and InstructBLIP \cite{dai2023instructblip} use a Q-Former for visual-language alignment that transfers visual features into a fixed number of learnable embeddings (32 in BLIP-2), which is similar to Perceiver IO \cite{jaegle2022perceiver} and Flamingo \cite{alayrac2022flamingo}. 

The Q-Former architecture is especially significant for its generalizability in aligning several modalities. This architecture of using cross-attention to transfer features to a small number of learnable embeddings has been used in recent studies for aligning many different modalities including image \cite{bai2023qwenvl, dai2023instructblip}  video \cite{zhang2023videollama}, and 3D \cite{hong20233dllm}.

However, despite the increased usage and significance of the Q-Former, prior research on its sublayers and their importance in different tasks has been limited. Elucidating the importance of each sublayer for different visual reasoning tasks can assist in designing more efficient training methods with effective parameter allocation. Moreover, although PEFT methods have been successfully applied to efficiently train language models \cite{hu2021lora, he2021towards, houlsby2019parameter, lester2021power, li2021prefix} evaluating the effectiveness of PEFT on the Q-Former and visual language models also remains under-explored. 
These two areas are critical for advancing the efficiency of training multimodal language models. 

In this work, we evaluate the performance of PEFT on InstructBLIP with two benchmarks, ScienceQA \cite{lu2022learn} and IconQA \cite{lu2021iconqa}, that respectively evaluate knowledge-grounded visual reasoning and perceptual visual reasoning. We apply LoRA to the Q-Former and base LLMs, Flan-T5-XL \cite{chung2022scaling} and Vicuna-7B \cite{vicuna2023}, and comprehensively test the performance of LoRA applied to different sublayers in the Q-Former with different ranks. We also examine the importance of each sublayer in the Q-Former for 4 different visual reasoning benchmarks using AdaLoRA \cite{zhang2023adalora}, which dynamically allocates parameter budgets to improve performance. To the best of our knowledge, we are the first to inspect the effectiveness of PEFT methods on the Q-Former and analyze its sublayers for visual reasoning tasks. 

Our contributions can be summarized as follows:
(1) We demonstrate that applying PEFT to the Q-Former can reduce the trainable parameters to less than 2\% while maintaining comparable performances.
(2) We show that in contrast to full fine-tuning the Q-Former and freezing the LLM, applying PEFT to both components can achieve superior results and reduces the total trainable parameters to less than 12\%.
(3) We examine the significance of the different sublayers in the Q-Former using AdaLoRA, and find that the self-attention layers are relatively more important for tasks that require stronger visual-language alignment, and more intrinsic ranks on FFN layers are needed to train on complex visual-language patterns.

\begin{figure}[t]
\centering
\includegraphics[width=1.0\columnwidth]{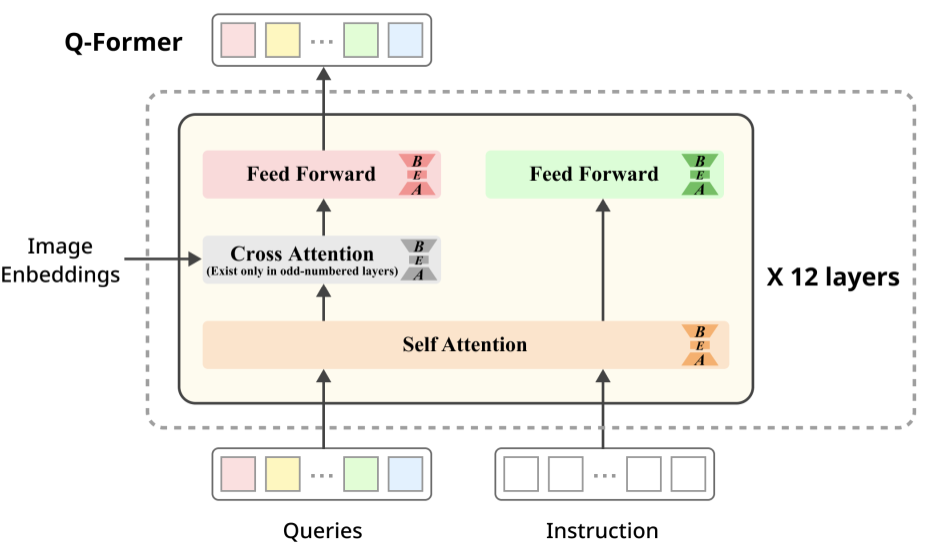}
\caption[]{The detailed structure of the Q-Former with AdaLoRA weight matrices (B, E, A).}
\label{fig:detailed_qformer}
\end{figure}

\begin{figure*}[t]
    \centering
    \begin{subfigure}{0.45\textwidth} 
        \includegraphics[width=\textwidth]{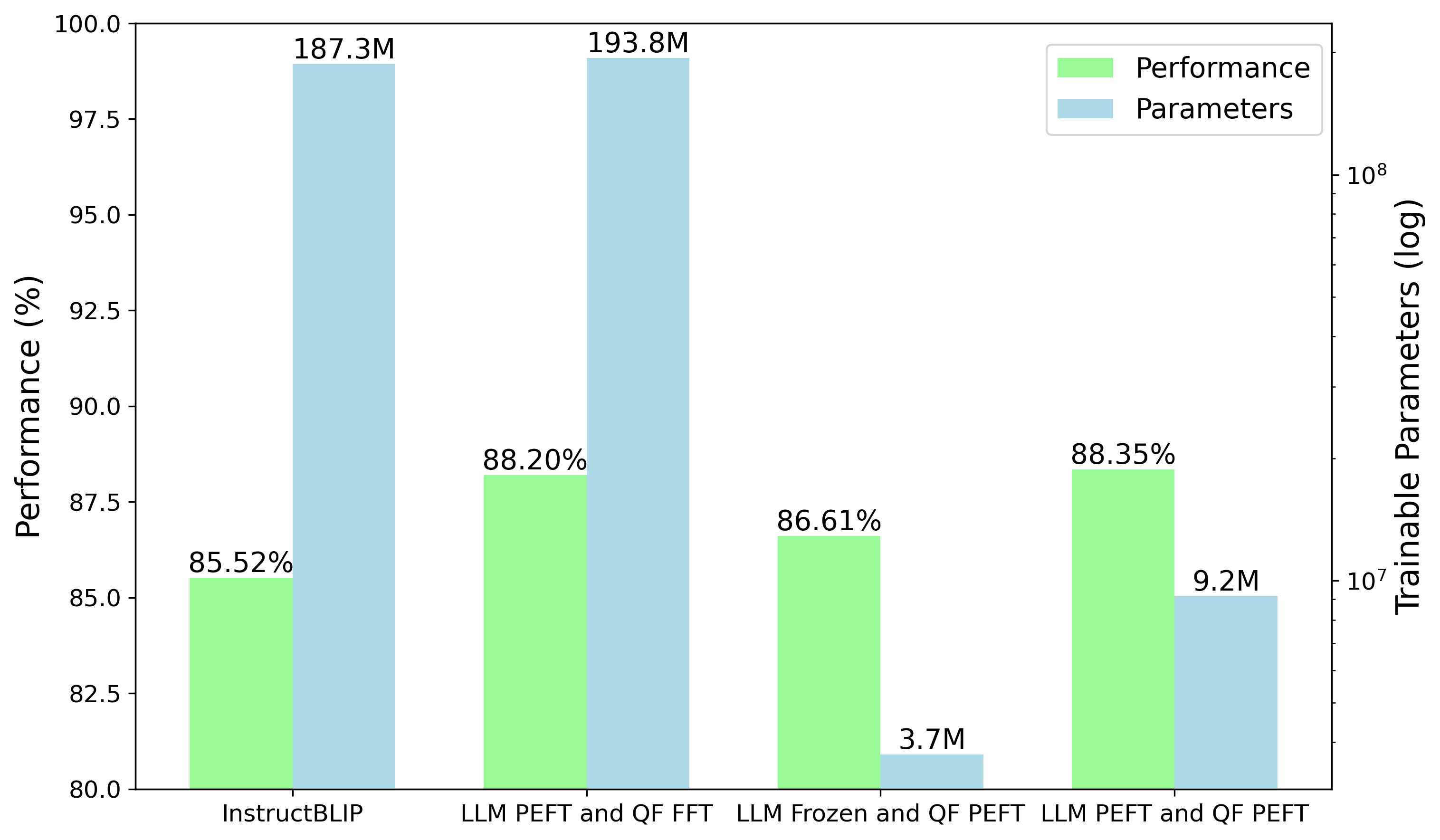}
        \caption{ScienceQA Flan-T5-XL}
        \label{fig:subfig1}
    \end{subfigure}
    \hfill  
    \begin{subfigure}{0.45\textwidth}  
        \includegraphics[width=\textwidth]{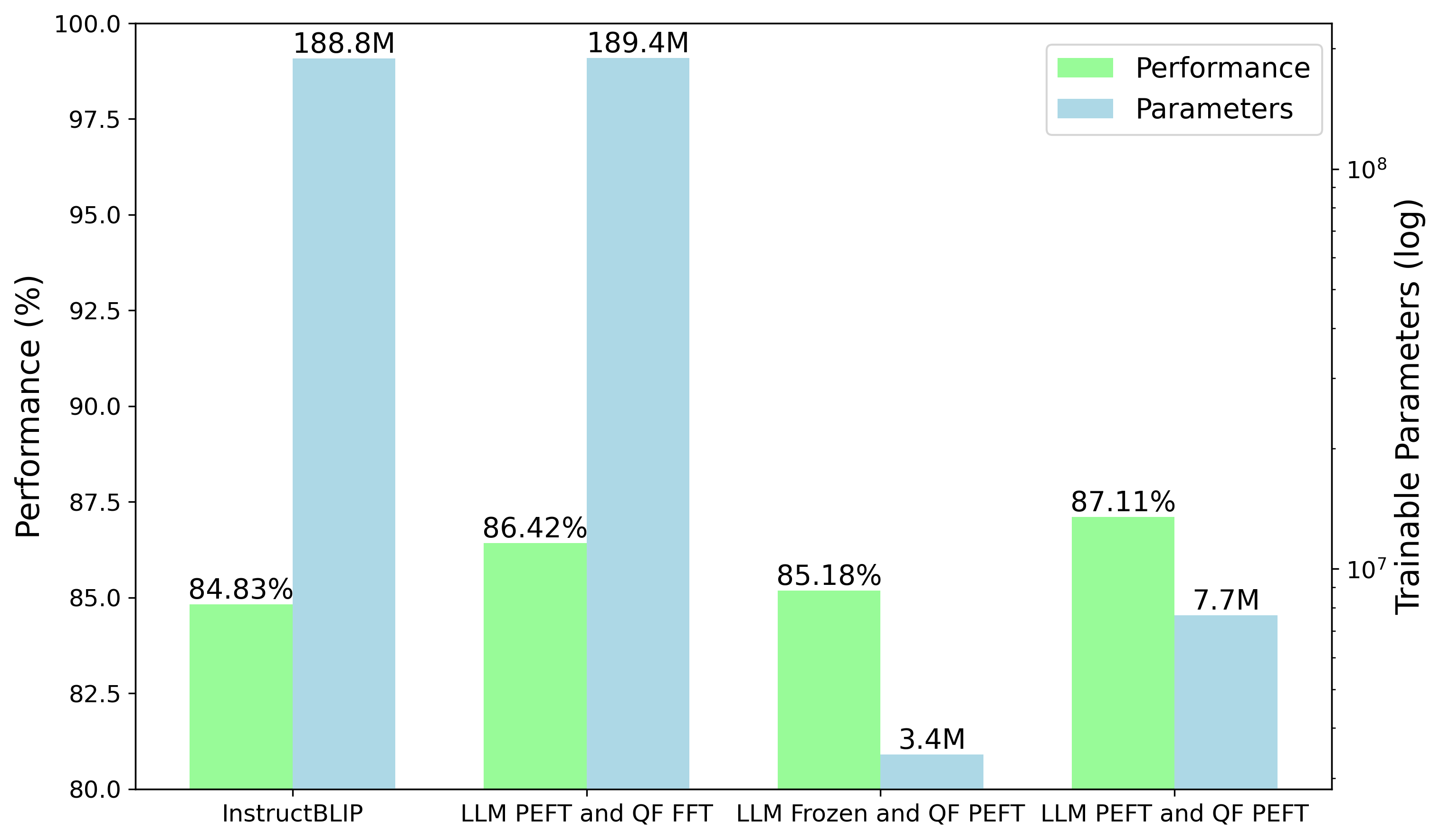}
        \caption{ScienceQA Vicuna-7B}
        \label{fig:subfig2}
    \end{subfigure}
    
    \vskip\baselineskip  
    \begin{subfigure}{0.45\textwidth}
        \includegraphics[width=\textwidth]{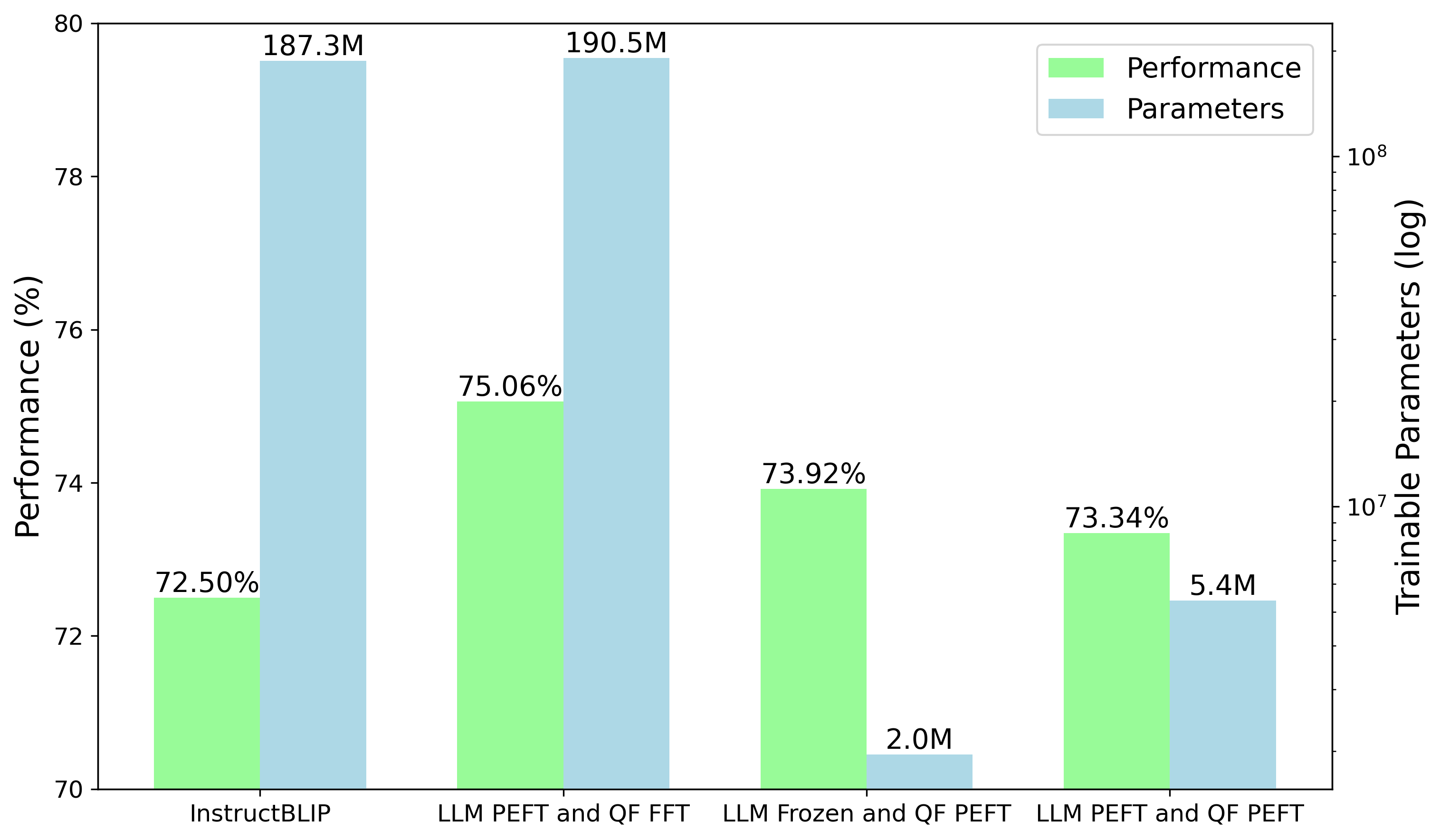}
        \caption{IconQA Flan-T5-XL}
        \label{fig:subfig3}
    \end{subfigure}
    \hfill
    \begin{subfigure}{0.45\textwidth}
        \includegraphics[width=\textwidth]{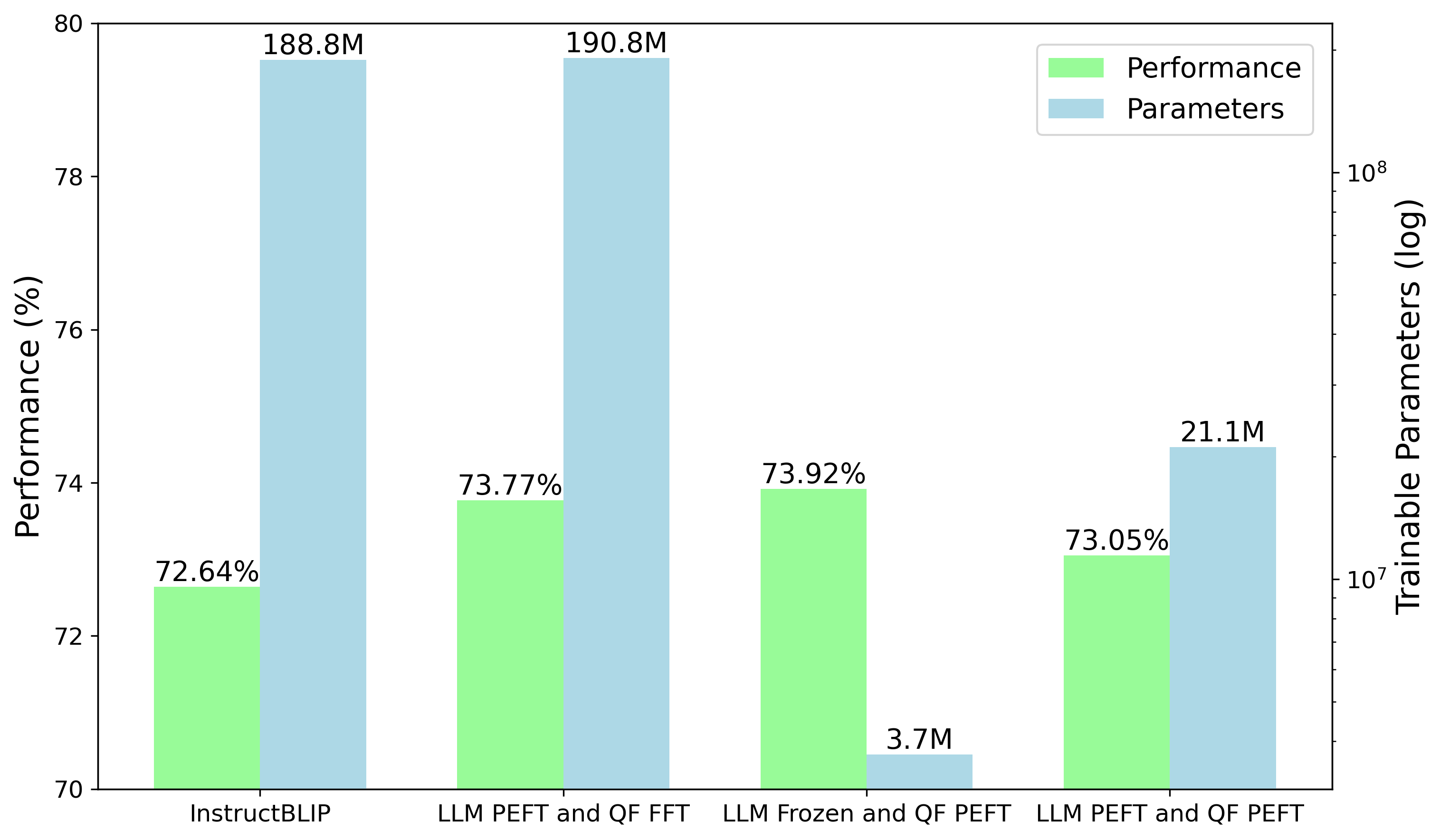}
        \caption{IconQA Vicuna-7B}
        \label{fig:subfig4}
    \end{subfigure}
    \caption{Comparing the performance and number of trainable parameters using Flan-T5-XL and Vicuna-7B as base models on ScienceQA and IconQA benchmarks. This compares the best performing configurations (rank value and LoRA-applied sublayers) of Q-Former full fine-tuning with LLM PEFT, Q-Former PEFT with frozen LLM, and Q-Former PEFT with LLM PEFT, against InstructBLIP (Q-Former full fine-tuning with frozen LLM). "QF" denotes Q-Former. "FFT" denotes full fine-tuning. The complete results and the training architectures are at Appendix~\ref{appendix-a}.}
    \label{fig:all_figs}
\end{figure*}

\section{Method}
\label{gen_inst}
In this work, we apply the PEFT method LoRA to the Q-Former and the LLM in InstructBLIP, and evaluate the performance on two visual reasoning benchmarks ScienceQA and IconQA. Additionally, we apply AdaLoRA to analyze the significance of each sublayer of the Q-Former on visual reasoning. 

\textbf{LoRA} reduces trainable parameters by decomposing the weight update matrix $\Delta W = BA$. After fine-tuning, the weight matrix can be reparametrized by adding the weight update to the original pre-trained model weights: $W + \Delta W = W + BA$, where $W \in \mathbb{R}^{d \times k}$, $B \in \mathbb{R}^{d \times r}$, $A \in \mathbb{R}^{r \times k}$, $r \ll \min(d, k)$. Unlike the original LoRA implementation, which confines its application to only the self-attention layers \cite{hu2021lora}, we extend the use of LoRA to multiple transformer sublayers in both the Q-Former and the LLM. Specifically, we apply LoRA to the q, v layers in self-attention layers, the q, k, v, o layers in cross-attention layers, and the FFN layers in the Q-Former. 

\renewcommand*{\thefootnote}{\arabic{footnote}}

\textbf{AdaLoRA} decomposes the intrinsic weight update matrix $\Delta W = BEA$
with singular value decomposition (SVD). During training, the less significant singular values are adaptively pruned based on their importance scores, adjusting the rank of the weight update matrices. The importance score of the $i$th singular value is calculated as follows: 
\begin{equation}
S_{i} = s(\lambda_{i}) + \frac{1}{d_1} \sum_{j=1}^{d_1} s(B_{ji}) + \frac{1}{d_2} \sum_{j=1}^{d_2} s(A_{ji})
\end{equation}
where $B \in \mathbb{R}^{d_1 \times r}$, $A \in \mathbb{R}^{r \times d_2}$ and $s(\cdot)$ is a specific importance function for each entry, based on sensitivity of each weight to the training loss.
As a result, applying AdaLoRA leads to appropriate rank allocation across modules for better performance. 
Since high-rank updates learn more complex signals, we use AdaLoRA for examining which sublayers in the Q-Former are critical for each visual reasoning task and which sublayers should be prioritized in parameter budget allocation for efficient fine-tuning. We apply AdaLoRA to the self-attention(q, v), cross-attention(q, k, v, o) layers, and FFN layers altogether for overall comparison. (Figure~\ref{fig:detailed_qformer})

\textbf{Base Models and Benchmarks.} We employ InstructBLIP as the base model for its pioneering use of the Q-Former and its strong performance on several downstream tasks \cite{dai2023instructblip} including ScienceQA (IMG) \cite{lu2022learn}, OCR-VQA \cite{8978122}, and A-OKVQA \cite{schwenk2022aokvqa}. We use the InstructBLIP implementation of LAVIS \cite{li-etal-2023-lavis} and use pre-trained Flan-T5-XL\footnote{\url{https://huggingface.co/google/flan-t5-xl}} and Vicuna-7B\footnote{\url{https://huggingface.co/lmsys/vicuna-7b-v1.3}} HuggingFace checkpoints in our experiments.

We use two benchmarks covering Knowledge Grounded Visual Reasoning (ScienceQA) and Perceptual Visual Reasoning (IconQA) \cite{lu2021iconqa} tasks. These benchmarks were held-out datasets for InstructBLIP, and were not involved in training the baseline InstructBLIP model. For analyzing the Q-Former with AdaLoRA we use two additional benchmarks, Vizwiz \cite{gurari2018vizwiz} and Flickr30k \cite{plummer2016flickr30k}.

\textbf{Knowledge Grounded Visual Reasoning} is a task of answering questions with a provided image related to the knowledge in diverse academic areas including physics, biology, and math. We use the ScienceQA dataset which covers a variety of science topics with corresponding extensive explanations. We only use the questions with image context (IMG). ScienceQA (IMG) has 6,218/2,097/2,017 samples for train/validation/test set.

\textbf{Perceptual Visual Reasoning} is a task of answering questions after comprehending the abstract meanings from an image. We use IconQA (Multi-text-choice) which contains question-answer pairs for natural images that require comprehensive reasoning abilities to understand abstract diagrams. IconQA (Multi-text-choice) has 18,946/6,316/6,316 samples for train/validation/test set.

\begin{figure*}[t]
    \centering
    \begin{subfigure}{0.48\textwidth} 
        \includegraphics[width=\textwidth]{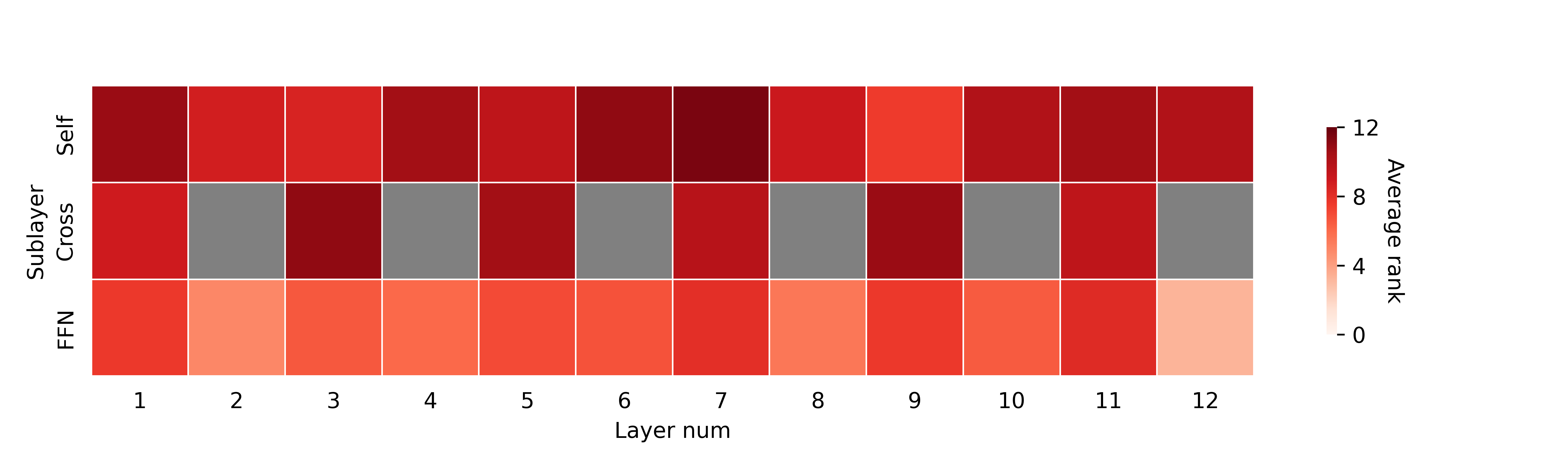}
        \caption{ScienceQA Flan-T5-XL}
        \label{fig:flansqa}
    \end{subfigure}
    \hfill
    \begin{subfigure}{0.48\textwidth} 
        \includegraphics[width=\textwidth]{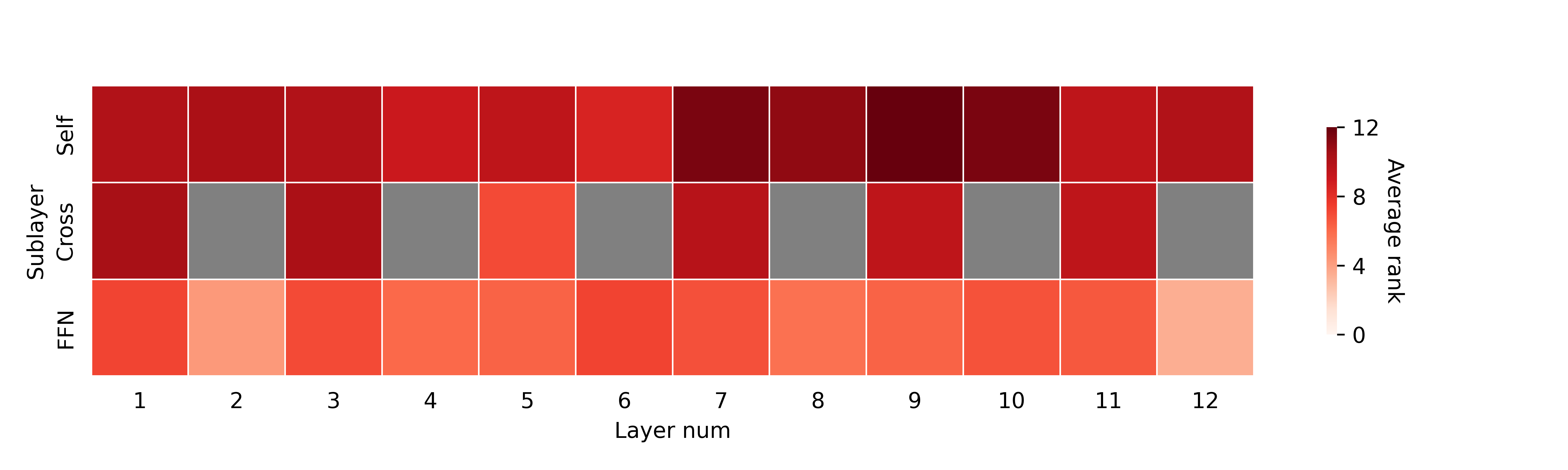}
        \caption{ScienceQA Vicuna-7B}
        \label{fig:vicunasqa}
    \end{subfigure}
    
    \vskip\baselineskip
    
    \begin{subfigure}{0.48\textwidth}
        \includegraphics[width=\textwidth]{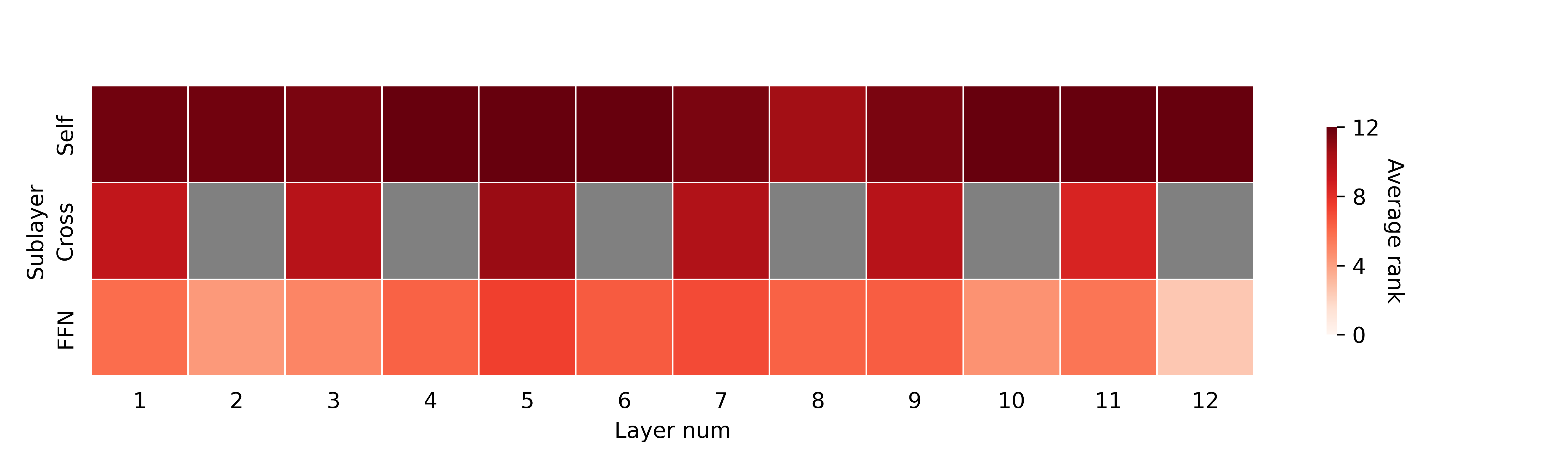}
        \caption{IconQA Flan-T5-XL}
        \label{fig:flanicon}
    \end{subfigure}
    \hfill
    \begin{subfigure}{0.48\textwidth}
        \includegraphics[width=\textwidth]{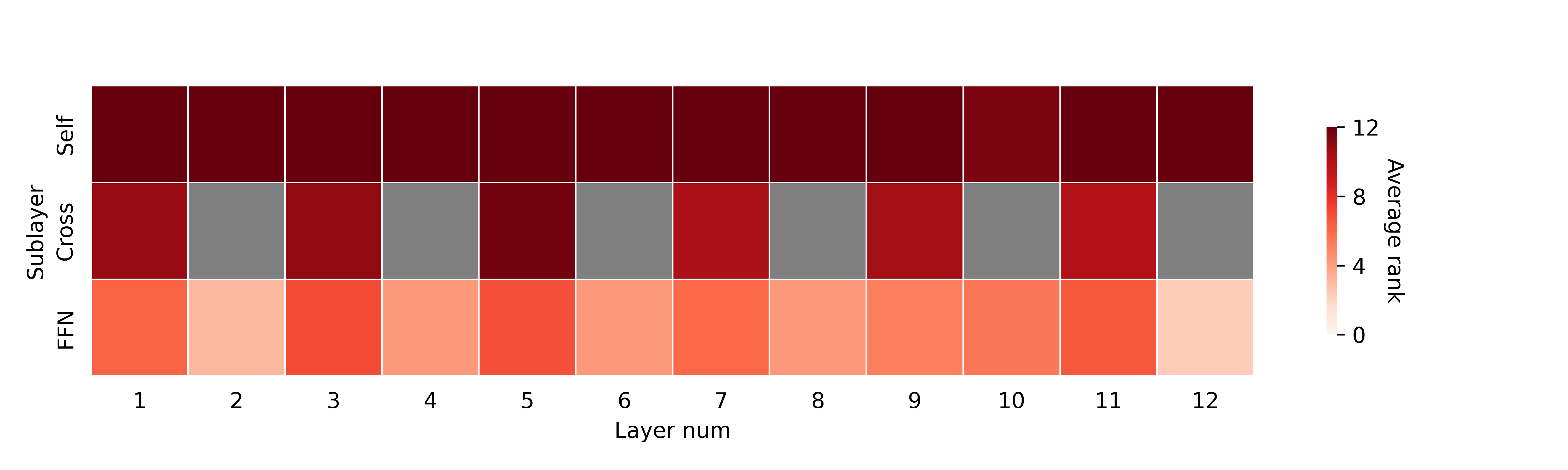}
        \caption{IconQA Vicuna-7B}
        \label{fig:vicunaicon}
    \end{subfigure}
    \caption{Heatmaps of the rank distributions of the sublayers in the Q-Former. Cross-attention layers are present in odd numbered layers only. Each value is the average of the component layers. The detailed heatmaps including additional benchmarks (Flickr30k, Vizwiz) are in Appendix~\ref{appendix-e}.}
    \label{fig:heatmap}
\end{figure*}

\section{Experiments}

\subsection{PEFT Effectiveness for Visual Reasoning}
We empirically analyze the effectiveness of training the Q-Former and the LLM in InstructBLIP with LoRA. (1) First, we apply LoRA to the LLM while still full fine-tuning the Q-Former. (2) Second, we apply LoRA to the Q-Former while freezing the LLM, resulting in efficient fine-tuning of the Q-Former. (3) Finally, we apply LoRA to both the Q-Former and the LLM. The performance comparison between (1), (2), (3) and the original InstructBLIP is in Figure~\ref{fig:all_figs}. (The main results of the overall experiments are in Appendix~\ref{appendix-a}, and the implementation and training details can be found in Appendix~\ref{appendix-b}.)

We find that applying LoRA to the Q-Former yields competitive performance, matching or surpassing full-fine-tuning while using less than 2\% of the original trainable parameters. Fine-tuning the base LLMs with LoRA consistently outperforms the baseline InstructBLIP model on both benchmarks, underscoring the enhanced task-specific language capabilities by training the language model. Applying LoRA to both the Q-Former and LLM achieve superior performance on both benchmarks with fewer than 12\% of the trainable parameters. Notably, we find that fine-tuning both models perform consistently higher in ScienceQA than in IconQA.
This discrepancy can be attributed to ScienceQA's richer language context. Given that ScienceQA entails more language information than IconQA, training the language model appears to yield a greater boost in performance. 

\subsection{Analysis of Q-Former Sublayers using AdaLoRA}
To investigate the significance of each sublayer in the Q-Former with 4 benchmarks (ScienceQA, IconQA, Flickr30k, Vizwiz), we use AdaLoRA to analyze the dynamically reallocated intrinsic ranks of the weight update matrices. We apply iteration-based AdaLoRA, with an initial rank of 12 and target rank of 8, to each training epoch. The final rank of each sublayer after training indicates their respective importance and the prioritization of the parameter budget. To visually represent these dynamics, we compute heatmaps (Figure~\ref{fig:heatmap}) that average AdaLoRA ranks within the same sublayers (self-attention, cross-attention, and FFN) across each of the Q-Former's 12 layers. These heatmaps illustrate the rank distribution for each training configuration across layers and sublayers.

For IconQA, rank allocation is predominantly focused on the self-attention layers for both base LLMs, with the cross-attention layers having the second highest number of ranks. Notably FFN layers tend to have higher ranks in odd-numbered layers, where the cross-attention layers are present. For ScienceQA, the FFN layers are similarly allocated the fewest average ranks. But compared to IconQA, the average rank distribution between the three sublayers are more balanced, and self-attention layers have noticeably fewer ranks. For all configurations, FFNs in the the final layer (12) consistently have the fewest ranks.

The distribution of rank allocation can be attributed to the different types of reasoning abilities required for each task and the relationship between sublayers. IconQA is consisted of perceptual visual reasoning questions that require strong visual-language alignment. Meanwhile, ScienceQA contains questions grounded in extensive knowledge, which demands significant logical reasoning on longer texts in addition to visual-language alignment.

Given that FFN layers are adept at learning task-specific patterns \cite{he2022unified}, and considering the complex textual patterns inherent in ScienceQA questions, we hypothesize that these factors contributed to the observed increase in ranks and influence of FFN layers for ScienceQA relative to IconQA. Also, FFN layers in odd-numbered layers tend to have higher ranks on both tasks, as they come after the cross-attention layers, receiving image features and making them more important for learning task-specific visual patterns. 

The rank allocation of self-attention layers can be attributed to the relative importance of visual-language alignment for the task. Self-attention layers allow query embeddings to attend textual information and extract visual features that are more relevant to the text prompt. This explains the significant concentration of ranks in the self-attention layers consistently throughout all 12 layers and base language models in IconQA. 

We use 2 additional benchmarks, Flickr30k and Vizwiz, for analyzing the Q-Former with AdaLoRA. Flickr30k is an image captioning task, and Vizwiz covers visual question answering.
The detailed results for each benchmark is shown in Appendix~\ref{appendix-e}.
The rank distribution is concentrated in the self-attention layers for both benchmarks, while the overall rank distribution is more even in Vizwiz than in Flickr30k. This result can be explained by the difference in complexity of the text prompts between the two benchmarks. Flickr30k’s text instruction is fixed to image captioning, while Vizwiz’s text instruction covers more diverse questions. Therefore, it can be explained that the resulting heatmap of Vizwiz aligns more with ScienceQA, and the result of Flickr30k aligns more with IconQA. This also indicates that result of AdaLoRA analysis in ScienceQA and IconQA generalizes well to other benchmarks.

\section{Conclusion}
\label{headings}
In this work, we systematically evaluate the effectiveness of applying PEFT to the Q-Former and visual language models. Our results show that applying PEFT to the Q-Former achieves comparable performance to full fine-tuning while only utilizing less than 2\% of the trainable parameters. Additionally, we employ dynamic parameter budget allocation with AdaLoRA to analyze the significance of the Q-Former's sublayers for different visual reasoning tasks. Our findings reveal that the importance of FFN layers increases when visual-language pattern becomes more complex, and importance of self-attention layers increases as significance of visual-language alignment in task increases.

\section*{Limitations}
More recently, the Q-Former architecture has been used to align many different modalities beyond images and languages including 3D, depth, audio, and video. In this work, we focus on efficiently training and analyzing the Q-Former for image-text alignment and leave the study of other modalities to future works.

\section*{Ethics Statement}
The datasets used in this work is publicly released by CC BY-NC-SA license, so there is no copyright issue in this paper.

\section*{Acknowledgements}
This work was supported by National Research Foundation of Korea (NRF) grant (RS-2023-00280883, RS-2023-00222663),  New Faculty Startup Fund from Seoul National University,  National Super computing Center (KSC-2023-CRE-0176),  Artificial Intelligence Industry Center Agency, Google cloud platform research credits, Seoul National University’s Engineering Department, and the AI Research Group AttentionX. 


\clearpage

\appendix

\section{InstructBLIP PEFT Experiments}
\label{appendix-a}
\begin{figure*}[t]
    \centering
    \includegraphics[width=1.0\linewidth]{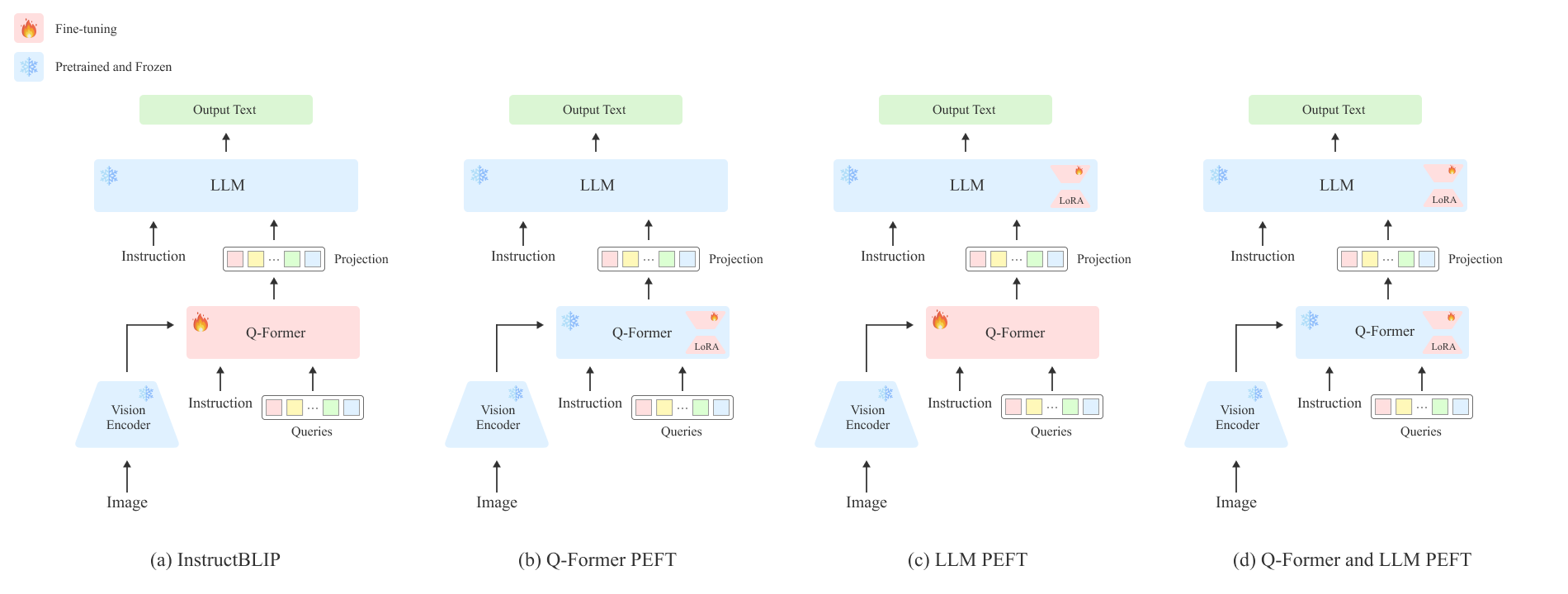} 
    \caption{Applying PEFT to the Q-Former and LLM in InstructBLIP.}
    \label{fig:model_architecture} 
\end{figure*}

The diagram of training configurations is shown in Figure~\ref{fig:model_architecture}. The full experimental results of applying PEFT to InstructBLIP, are shown in Table~\ref{table:main_table}. All the results presented in this paper are obtained after single-run experiments.

\section{Model Training Details for PEFT Evaluations}
\label{appendix-b}
We conduct each experiment in Table~\ref{table:main_table} and Figure~\ref{fig:all_figs} using a single A100 GPU. We set the maximum epoch to 15 with early stopping of 3 patience steps. We use linear decay as a learning rate scheduler with the AdamW optimizer. For the initial learning rate, we primarily use 2e-5 for experiments which involves full fine-tuning the Q-Former, and otherwise 5e-4. For certain cases that deviate significantly from other experiments, we lower the learning rate from 2e-5 to 1e-5 and 5e-4 to 1e-4. These cases include: (1) When the model is trained on less than 8 epochs (the halfway point) by early stopping, (2) When the training is considered unstable, i.e. resulting in over 10\%p lower performance than other experiment in an equivalent setup having different r value. We set the weight decay to 0.05. For batch size, we use 16 as an effective batch size across all experiments. Only difference is that (batch size, gradient accumulation iterations) were set to (8, 2) for Vicuna-7B and (16, 1) for Flan-T5-XL.

\section{Model Training Details for AdaLoRA Experiments}
\label{appendix-c}

For ScienceQA and IconQA, epoch settings, effective batch sizes (16), learning rate and scheduling methods, and weight decay values are given the same as Appendix~\ref{appendix-b}. For Flickr30k and Vizwiz, we set the maximum epoch to 5 with early stopping of 3 patience steps. We use “linear\_warmup\_cosine\_lr” scheduler, and set an initial learning rate of 1e-4 with batch size 8 on Vizwiz, and set an initial learning rate of 5e-5 with batch size 60 on Flickr30k.
\section{Instruction Templates}
\label{appendix-d}
We provide instructions used in ScienceQA and IconQA. We use the same format from the InstructBLIP paper. We add alphabet labels for each choices and the answer. For ScienceQA, we construct the "context" section of the instruction by incorporating information from both the 'hint' and 'lecture' fields, if they are available in the dataset.

\vspace{1em}
\textbf{ScienceQA}
Context: \{ \{{hint}\} \{lecture\} \} Question: \{ \{question\} \} Options: \{ \{choices\} \}. Answer:

\textbf{IconQA}
<Image> Question: \{ \{question\} \}  Options: \{ \{choices\} \}. Short answer:
\vspace{0pt}

\begin{figure}[H]
\centering
\includegraphics[width=0.4\textwidth]{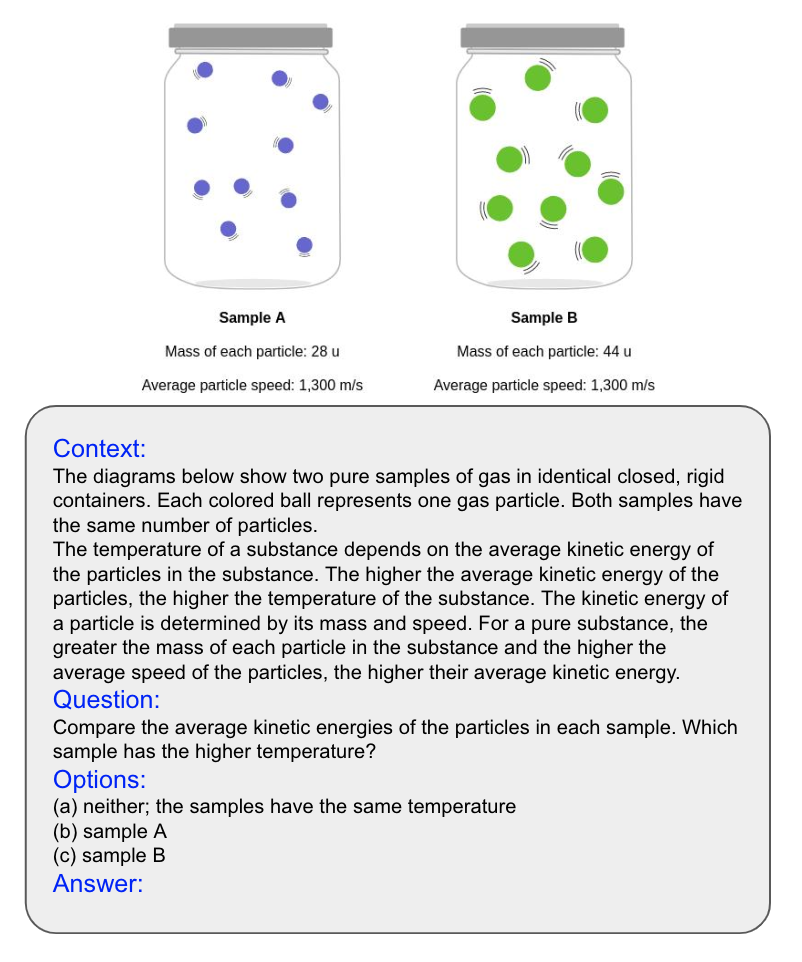}
\caption[Example ScienceQA instruction template]{Example ScienceQA\footnotemark[3] instruction template}
\label{fig:scienceqa_example_template}
\end{figure}

\begin{figure}[H]
\centering
\includegraphics[width=0.4\textwidth]{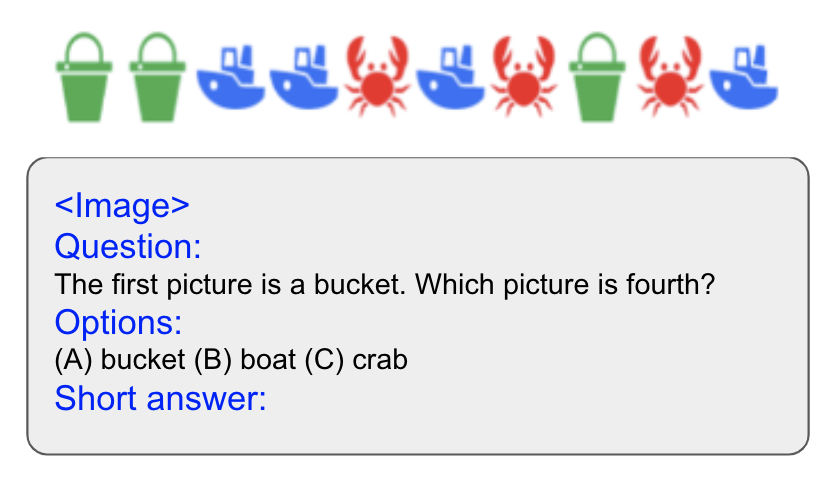}
\caption[Example IconQA instruction template]{Example IconQA\footnotemark[3] instruction template}
\label{fig:iconqa_example_template}
\end{figure}

\footnotetext[3]{\url{https://creativecommons.org/licenses/by-nc-sa/4.0/}}
\section{Detailed Figures for AdaLoRA Experiments}
\label{appendix-e}
\begin{figure*}[t]
    \centering
    \begin{subfigure}{0.8\textwidth}
        \includegraphics[width=\textwidth]{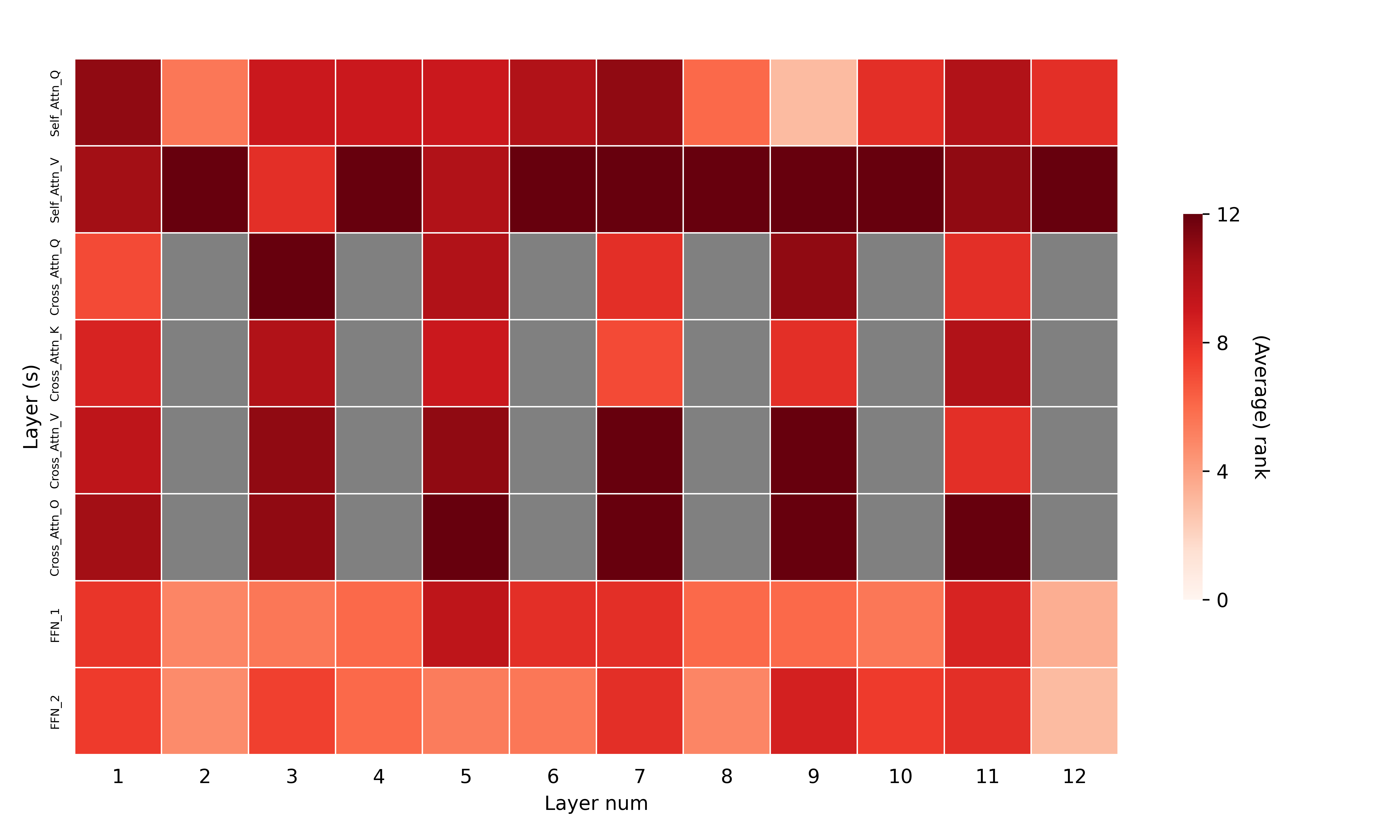}
        \caption{ScienceQA Flan-T5-XL}
        \label{fig:flansqa_scf}
    \end{subfigure}
    \vfill 
    \begin{subfigure}{0.8\textwidth} 
        \includegraphics[width=\textwidth]{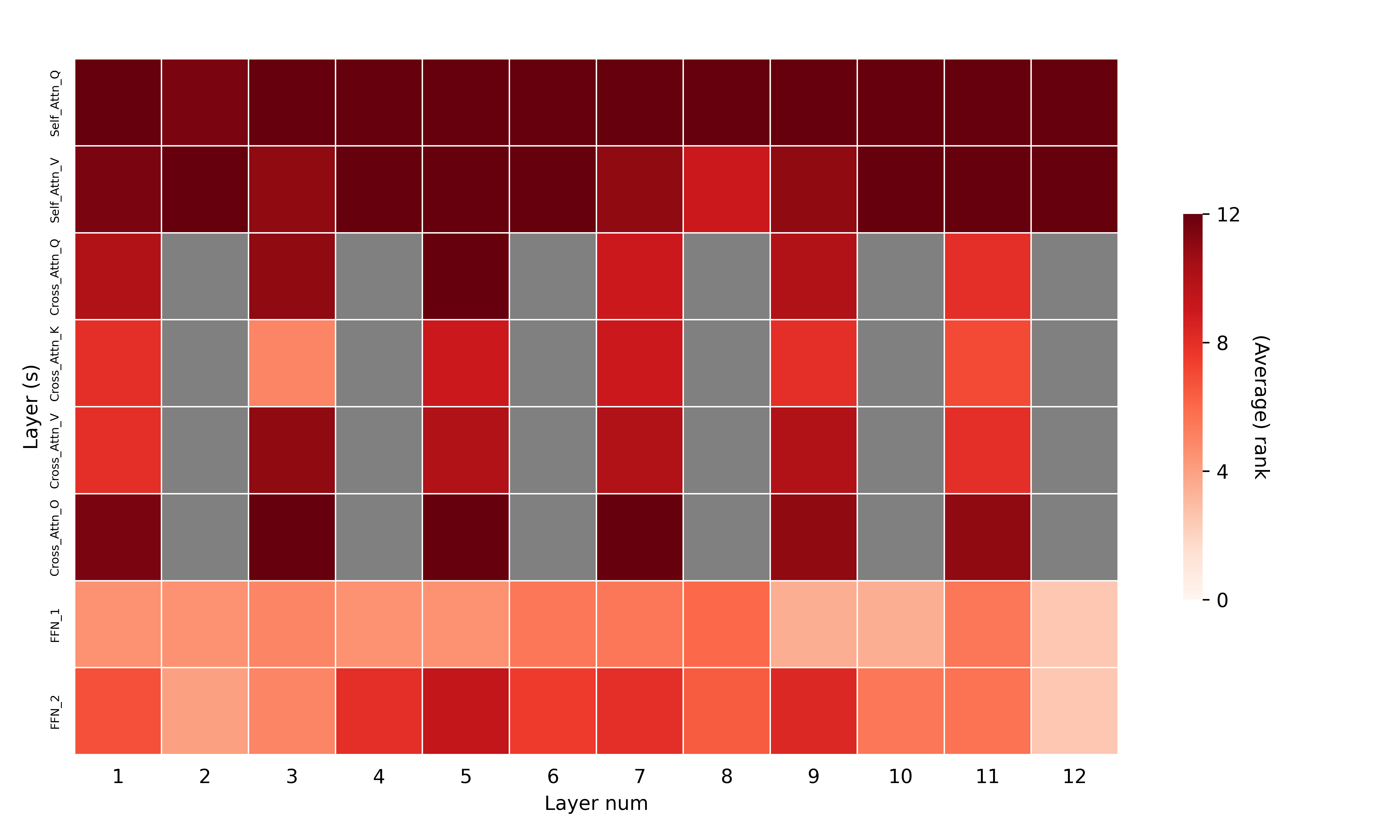}
        \caption{IconQA Flan-T5-XL}
        \label{fig:flanicon_scf}
    \end{subfigure}
    \caption{Detailed heatmaps of rank distribution of modules in layers of the Q-Former. (Flan-T5-XL as a base LLM) Cross-attention layers are present in odd numbered layers only.  The rank values in the feed-forward network (FFN) components are averaged across both FFN layers. }
    \label{fig:heatmap6}
\end{figure*}

\begin{figure*}[t]
    \centering
    \begin{subfigure}{0.8\textwidth}
        \includegraphics[width=\textwidth]{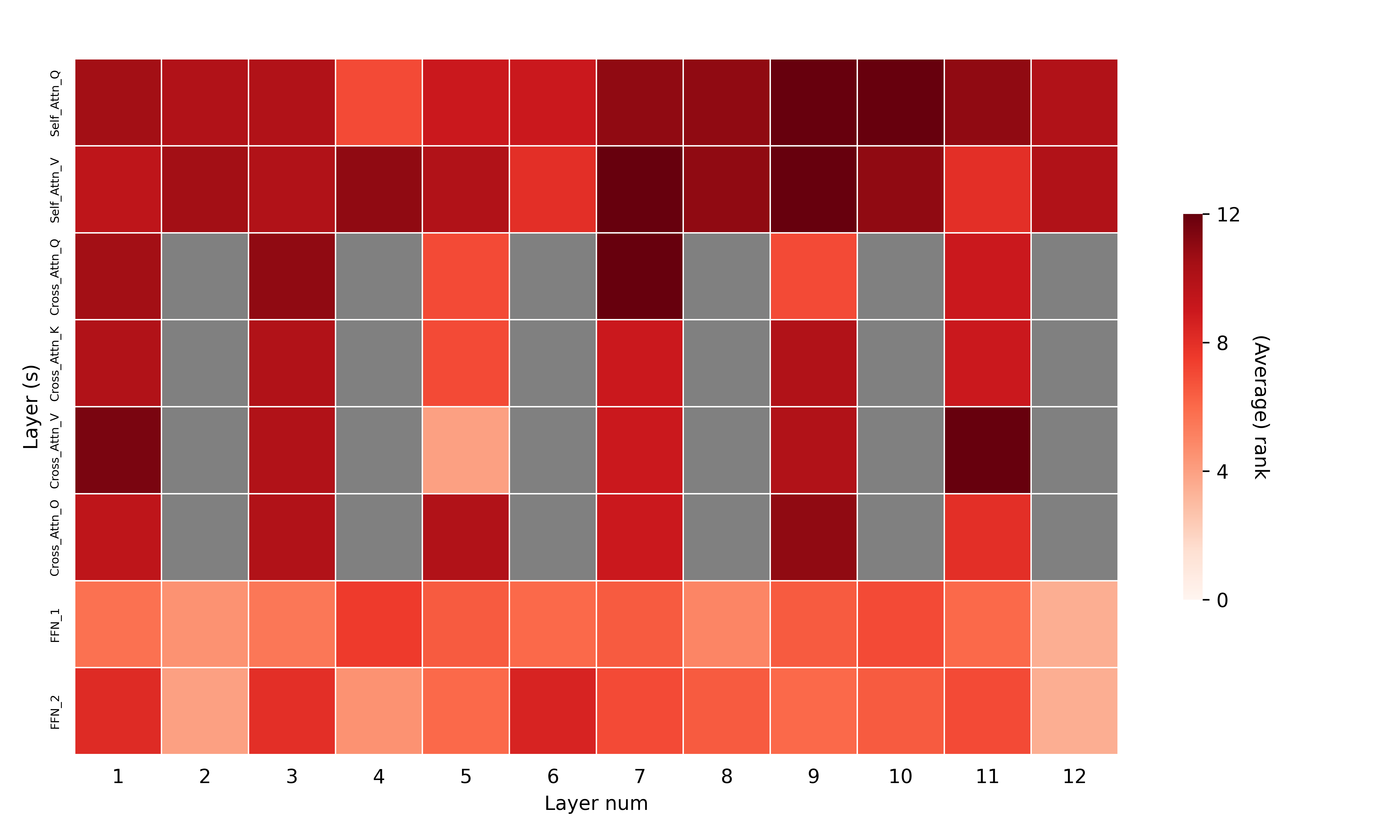}
        \caption{ScienceQA Vicuna-7B}
        \label{fig:vicunasqa_scf}
    \end{subfigure}
    \vfill 
    \begin{subfigure}{0.8\textwidth}
        \includegraphics[width=\textwidth]{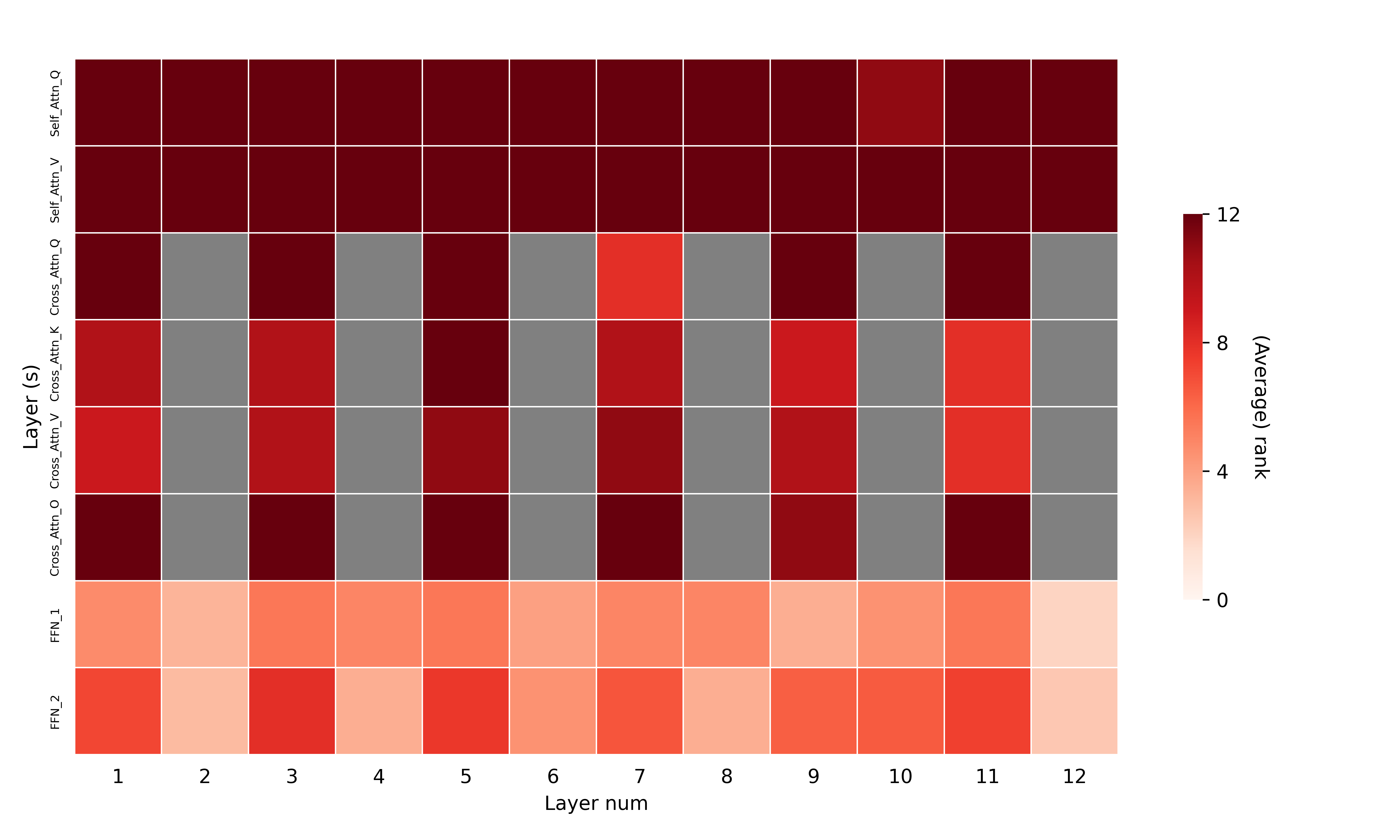}
        \caption{IconQA Vicuna-7B}
        \label{fig:vicunaicon_scf}
    \end{subfigure}
    
    \caption{Detailed heatmap of rank distribution of modules in layers of the Q-Former. (Vicuna-7B as a base LLM) Cross-attention layers are present in odd numbered layers only.  The rank values in the feed-forward neural network (FFN) components are averaged across both FFN layers.}
    \label{fig:heatmap_detailed_vicuna}
\end{figure*}

\begin{figure*}[t]
    \centering
    \begin{subfigure}{0.48\textwidth}
        \includegraphics[width=\textwidth]{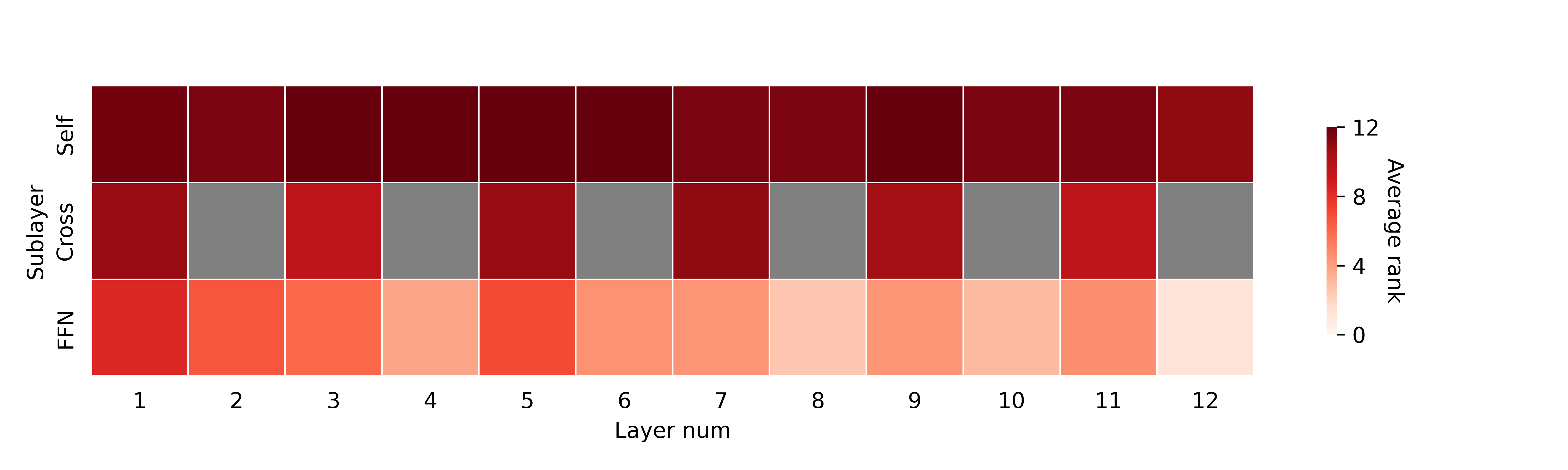}
        \caption{Flickr30k Flan-T5-XL}
        \label{fig:flanflickr_scf_summarized}
    \end{subfigure}
        \begin{subfigure}{0.48\textwidth}
        \includegraphics[width=\textwidth]{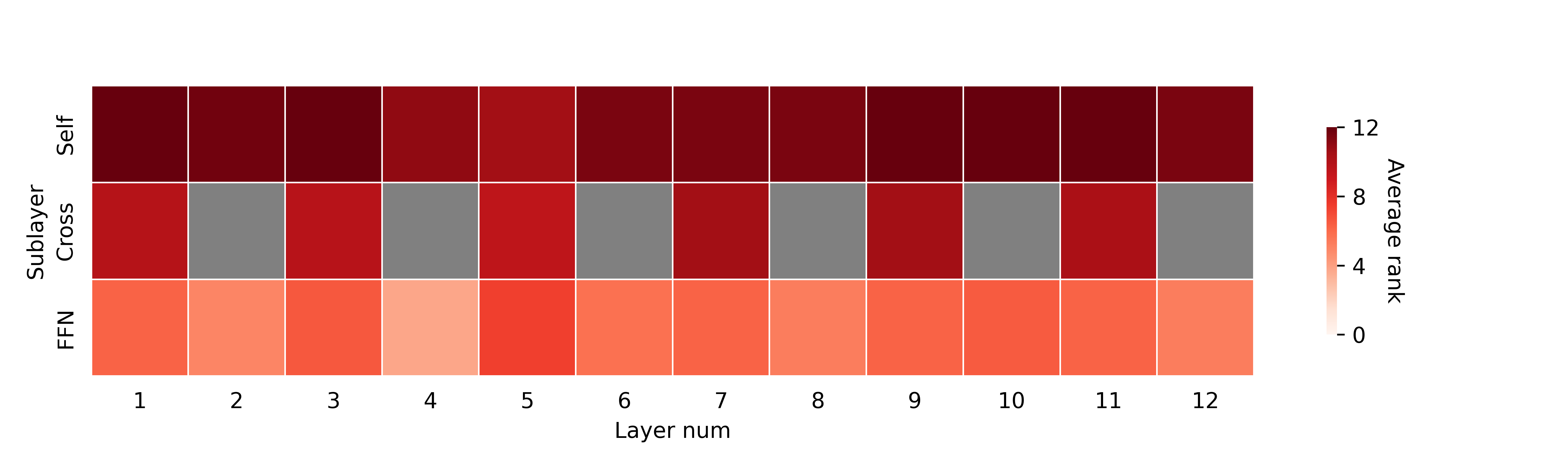}
        \caption{Vizwiz Flan-T5-XL}
        \label{fig:flanviz_scf_summarized}
    \end{subfigure}
    \caption{Heatmaps of the rank distributions of the sublayers in the Q-Former for Flickr30k and Vizwiz. (Flan-T5-XL as a base LLM) Cross-attention layers are present in odd numbered layers only. Each value is the average of the component layers.}
    \label{fig:heatmap_summarized_flan_flickr_vizwiz}
\end{figure*}

\begin{figure*}[t]
    \centering
    \begin{subfigure}{0.8\textwidth}
        \includegraphics[width=\textwidth]{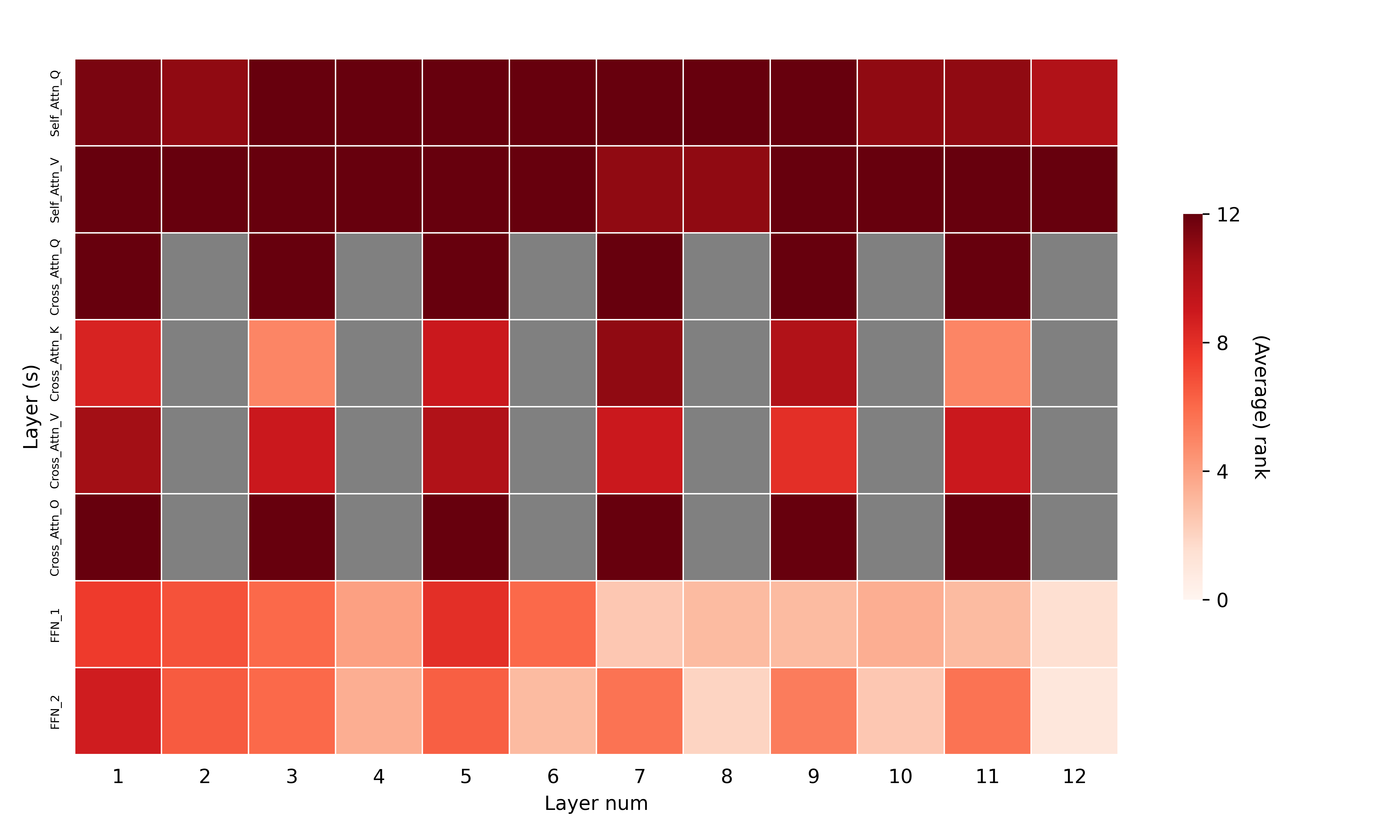}
        \caption{Flickr30k Flan-T5-XL}
        \label{fig:flanflickr_scf}
    \end{subfigure}
    \vfill 
    \begin{subfigure}{0.8\textwidth}
        \includegraphics[width=\textwidth]{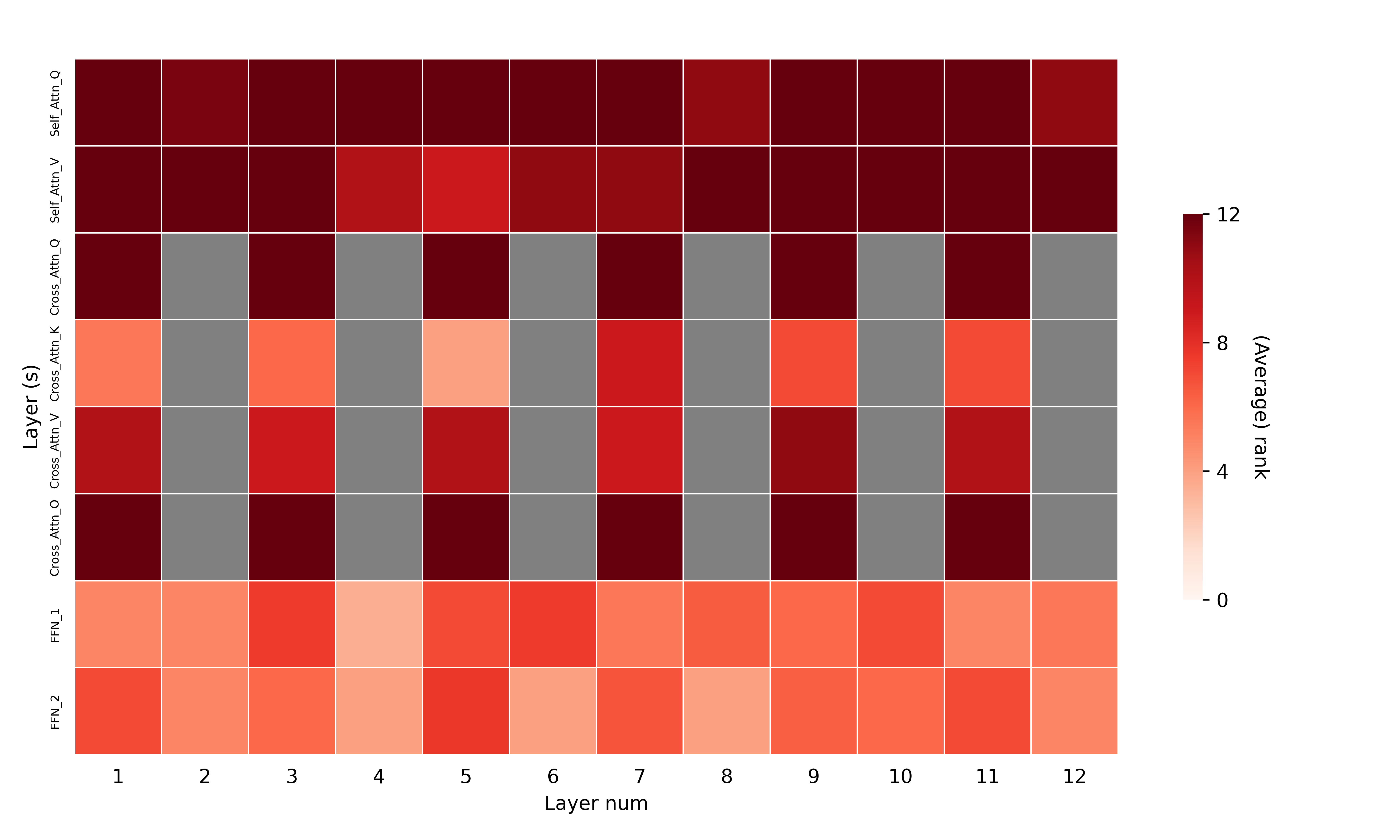}
        \caption{Vizwiz Flan-T5-XL}
        \label{fig:flanviz_scf}
    \end{subfigure}
    \caption{Detailed heatmaps of rank distribution of modules in layers of the Q-Former for Flickr30k and Vizwiz. (Flan-T5-XL as a base LLM) Cross-attention layers are present in odd numbered layers only.  The rank values in the feed-forward network (FFN) components are averaged across both FFN layers. }
    \label{fig:heatmap_detailed_flan_flickr_vizwiz}
\end{figure*}

The detailed figures of rank distribution in AdaLoRA experiments are shown in Figure~\ref{fig:heatmap6} (Flan-T5-XL) and Figure~\ref{fig:heatmap_detailed_vicuna} (Vicuna-7B). Also, heatmaps for Flickr30k, Vizwiz with Flan-T5-XL are shown in Figure~\ref{fig:heatmap_summarized_flan_flickr_vizwiz} and Figure~\ref{fig:heatmap_detailed_flan_flickr_vizwiz}.

\begin{table*}
    \centering 
    \begin{adjustbox}{width=\textwidth,center}
    \begin{tabular}{lcccccccccccc} 
        \toprule 
        & \multicolumn{4}{c}{Method} & \multicolumn{4}{c}{ScienceQA} & \multicolumn{4}{c}{IconQA} \\
        \cmidrule(lr){2-5} \cmidrule(lr){6-9} \cmidrule(lr){10-13}
        & LLM & Q-Former & Sublayer & Base Model & r=1 & r=2 & r=4 & r=8 & r=1 & r=2 & r=4 & r=8 \\
        \midrule
        & LoRA & Full & ffn & Flan-T5-XL & \textbf{86.42} & 86.37 & 85.32 & 86.27  & 73.40 & \textbf{74.76} & 74.00 & 71.52 \\
        & LoRA & Full & attn & Flan-T5-XL & \textbf{87.36} & 86.17 & 86.91 & 86.42 & 72.34 & 72.88 & \textbf{73.45} & 73.29 \\
        & LoRA & Full & all & Flan-T5-XL & 87.41 & 87.36 & \underline{\textbf{88.20}} & 87.90 & 72.61 & \underline{\textbf{75.06}} & 73.23 & 72.74 \\
        \specialrule{0.01pt}{1pt}{1pt}
        & Freeze & LoRA & ffn & Flan-T5-XL & 84.83 & 83.79 & 83.14 & \textbf{85.87} & 70.54 & 72.13 & 68.08 & \textbf{72.40} \\
        & Freeze & LoRA & self-attn & Flan-T5-XL & \textbf{86.02} & 83.74 & 79.57 & 86.02 & 71.82 & \textbf{72.55} & 72.06 & 71.64 \\
        & Freeze & LoRA & cross-attn & Flan-T5-XL & 84.13 & \textbf{86.32} & 84.88 & 85.18 & 72.32 & 72.42 & 72.32 & \underline{\textbf{73.92}} \\
        & Freeze & LoRA & all & Flan-T5-XL & 85.37 & 86.42 & 83.89 & \underline{\textbf{86.61}} & 70.19 & 70.50 & 72.82 & \textbf{73.31} \\
        \specialrule{0.01pt}{1pt}{1pt}
        & LoRA & LoRA & all & Flan-T5-XL & 88.00 & 88.10 & \underline{\textbf{88.35}} & 88.05 & 71.47 & \underline{\textbf{73.34}} & 71.41 & 73.18 \\
        \toprule
        & LoRA & Full & ffn & Vicuna-7B & 86.32 & \textbf{86.42} & 85.87 &  85.97 & 71.39 & 72.97 & \textbf{73.02} & 72.34 \\
        & LoRA & Full & attn & Vicuna-7B & \underline{\textbf{86.42}} & 86.32 & 85.08 & 85.23 & 72.36 & \textbf{73.16} & 72.29 & 73.02 \\
        & LoRA & Full & all & Vicuna-7B & 85.03 & \textbf{86.32} & 85.57 & 85.72 & \underline{\textbf{73.77}} & 71.71 & 72.93 & 73.15 \\
        \specialrule{0.01pt}{1pt}{1pt}
        & Freeze & LoRA & ffn & Vicuna-7B & 83.44 & \textbf{83.74} & 83.64 & 83.74 & 69.89 & \textbf{72.50} & 72.50 & 71.11 \\
        & Freeze & LoRA & self-attn & Vicuna-7B & \textbf{83.19} & 81.51 & 82.25 & 83.14 & 71.23 & 71.45 & 71.42 & \textbf{71.74} \\
        & Freeze & LoRA & cross-attn & Vicuna-7B & \textbf{83.29} & 83.24 & 83.14 & 82.75 & 71.11 & 72.40 & 71.99 & \textbf{73.39} \\
        & Freeze & LoRA & all & Vicuna-7B & \underline{\textbf{85.18}} & 82.80 & 83.74 & 83.44 & 71.49 & \underline{\textbf{73.92}} & 71.45 & 73.40 \\
        \specialrule{0.01pt}{1pt}{1pt}
        & LoRA & LoRA & all & Vicuna-7B  & 85.87 & \underline{\textbf{87.11}} & 85.08 & 85.62 & 71.72 & 72.01 & 72.61 & \underline{\textbf{73.05}}\\
        \bottomrule
    \end{tabular}
    \end{adjustbox}
    \caption{Overall performance results. "Full" indicates full fine-tuning, and the best results among 4 r values are bolded. The best results for each PEFT category, benchmark, and base language models are underlined. The underlined performances are used to compare the best performances between PEFT methods in Figure~\ref{fig:all_figs}.
    }
    \label{table:main_table}
\end{table*}

\vspace{-50pt}  

\end{document}